%% file: main.tex
\documentclass[10pt,twocolumn,letterpaper]{article}

\usepackage{cvpr}              
\input{preamble}
\definecolor{cvprblue}{rgb}{0.21,0.49,0.74}
\usepackage[pagebackref,breaklinks,colorlinks,allcolors=cvprblue]{hyperref}
\usepackage{pifont}    
\usepackage{amssymb}   
\usepackage{booktabs}  
\def\paperID{33} 
\def\confName{CVPR}
\def\confYear{2026}

\title{\tool: A Multi-View Annotation System for Multi-Object Tracking}

\author{
Hao Vo,
Duc Nguyen,
Ngan Le \\
AICV Lab, University of Arkansas, USA  \\[2pt]
{%
\ttfamily\small
\makebox[\linewidth][c]{%
\parbox{0.95\linewidth}{\centering
\tt\small \{haov,dnguyen3,thile\}@uark.edu
}}%
}%
}

\newcommand{\tool}{\textsc{ArgusTrack}\xspace}
\begin{document}
\maketitle
\input{sec/0_abstract}    
\input{sec/1_intro}
\input{sec/2_method}
\input{sec/3_exp}
{
    \small
    \bibliographystyle{ieeenat_fullname}
    \bibliography{main}
}

\end{document}

%% file: sec/0_abstract.tex
\begin{abstract}
Multi-Camera Multi-Target (MCMT) tracking has emerged as a critical capability for applications ranging from autonomous driving to animal behavior monitoring. While recent advances have yielded sophisticated tracking algorithms, the availability of annotated multi-view data remains a significant bottleneck. Existing annotation tools predominantly support single-camera workflows or rely on LiDAR sensors, making cross-view labeling tedious and impractical for camera-only setups. We present \tool, a multi-camera annotation system that addresses these limitations by enabling annotators to work directly on a bird's-eye-view (BEV) plane. Given calibrated camera parameters, a single ground-plane annotation is automatically projected into 2D bounding boxes across all relevant views, inherently ensuring identity consistency without manual cross-view alignment. To further accelerate the labeling process, \tool incorporates two complementary mechanisms: a Temporal Aware module that propagates annotations from preceding frames to initialize new ones, requiring only minor positional adjustments; and a Multi-camera Semi-annotation module that leverages off-the-shelf 2D detectors combined with foot-point estimation to automatically generate candidate BEV positions for annotator verification. We evaluate \tool through a pilot study on multi-camera broiler tracking and demonstrate that it substantially reduces annotation time compared to conventional single-camera labeling workflows. More details can be found on our project page at \url{https://uark-aicv.github.io/argustrack/}.
\end{abstract}

%% file: sec/1_intro.tex
\vspace{-6mm}

\section{Introduction}

Multi-camera systems have become increasingly important for animal behavior monitoring~\cite{cardoen2025multi, phan5333983broilertrack, cardoen2024label, aou2024multi, xu2025multi, yamamoto2025entire}, where frequent occlusions and high visual similarity among individuals pose significant challenges for single-camera setups. By observing the same scene from multiple viewpoints, multi-camera configurations can resolve ambiguities that arise when animals overlap or occlude one another, enabling more reliable tracking and identity preservation. This capability is particularly valuable in livestock monitoring~\cite{phan5333983broilertrack}, laboratory animal studies~\cite{yamamoto2025entire}, and wildlife observation~\cite{xu2025multi}, where accurate individual-level tracking is essential for behavioral analysis. Beyond animal monitoring, Multi-Camera Multi-Target (MCMT) tracking~\cite{li2019state, li2021visio, fan2025all, daryani2025camuvid, cheng2023rest} has broad applications in autonomous driving~\cite{ding2024ada, wang2023exploring}, surveillance~\cite{li2020infrared}, and crowd analysis~\cite{ma2019bayesian}.

\begin{figure}[t]
\centering
\includegraphics[width=\linewidth]{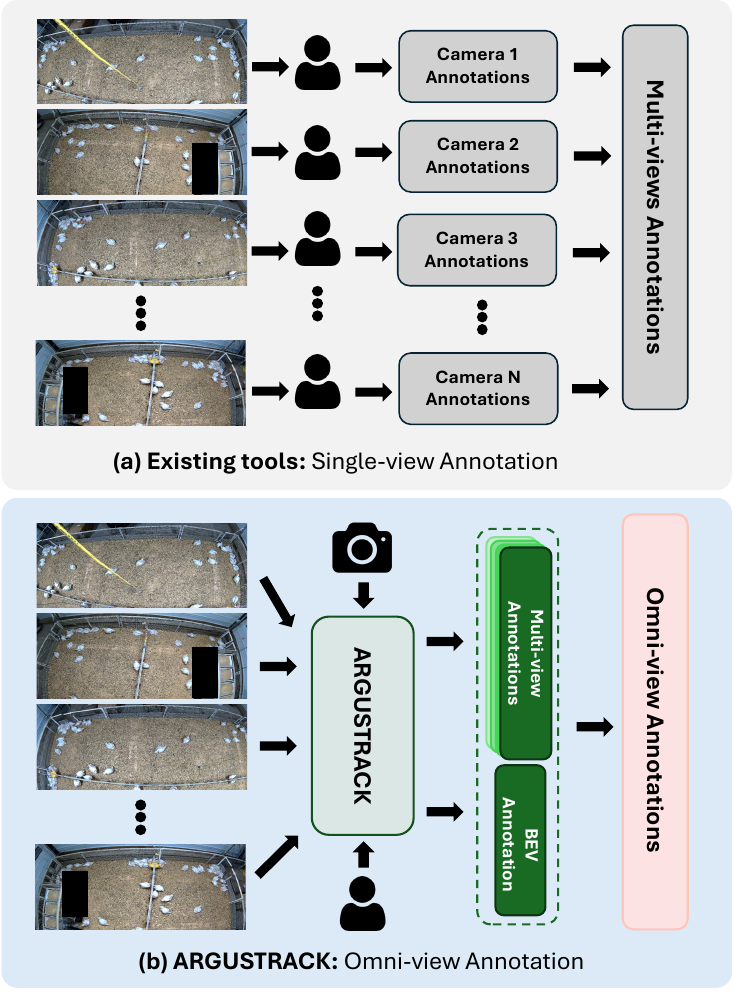}
\vspace{-6mm}

\caption{\textbf{Motivation.}
(a) Conventional single-camera annotation requires annotators to label each camera view independently.
(b) Our Omni-based tool lets annotators label targets once on the BEV plane, with annotations automatically projected to all camera views for consistent identity synchronization.}
\label{fig:multi_camera_motivation}
\vspace{-6mm}

\end{figure}
Recent years have witnessed notable progress in MCMT algorithms for animal tracking. BroilerTrack~\cite{phan5333983broilertrack} projects detections onto the bird's-eye-view (BEV) plane for spatial clustering, while other methods incorporate temporal cues~\cite{teepe2024lifting} or learned feature representations~\cite{teepe2024earlybird} to improve association accuracy. However, despite these algorithmic advances, the lack of annotated multi-camera data remains a critical bottleneck for training and evaluating such methods.

Existing annotation tools are ill-suited for multi-camera scenarios. Popular labeling platforms such as CVAT~\cite{CVAT_ai_Corporation_Computer_Vision_Annotation_2023}, Label Studio~\cite{Label_Studio}, and LabelMe~\cite{Wada_Labelme_Image_Polygonal} support only single-camera workflows, requiring annotators to label each view independently and manually ensure identity consistency across cameras---a tedious and error-prone process. Alternative tools designed for 3D annotation~\cite{lee2025openbox, zimmermann20193d} rely on LiDAR sensors, which are rarely deployed in animal monitoring environments. As a result, researchers often resort to laborious manual procedures that do not scale (Figure~\ref{fig:multi_camera_motivation}a). Table~\ref{tab:annotation_comparison} highlights the limitations of single-camera annotation workflows and the advantages of our multi-camera approach.

\begin{table}[t]
\centering
\caption{Comparison of annotation tools for multi-camera tracking.
Single-view tools require independent annotation per camera, causing
effort to grow with the number of cameras, whereas our tool annotates
in an omni-view space with constant effort regardless of camera count.}
\vspace{-2mm}
\label{tab:annotation_comparison}
\resizebox{0.9\linewidth}{!}{%
\begin{tabular}{l c c c c}
\toprule
\textbf{Tool} 
& \textbf{\shortstack{Annotation\\Space}} 
& \textbf{\shortstack{Annotation\\Time}} 
& \textbf{\shortstack{Cross-view\\Consistency}} 
& \textbf{\shortstack{Multi-cam\\Scalability}} \\
\midrule
CVAT \cite{CVAT_ai_Corporation_Computer_Vision_Annotation_2023}
& Single-view 
& High 
& Manual 
& Limited \\
TrackMe \cite{phan2024trackme}
& Single-view 
& High 
& Manual 
& Limited \\
LabelMe \cite{Wada_Labelme_Image_Polygonal}
& Single-view 
& High 
& Manual 
& Limited \\
\midrule
\tool~(Ours) 
& \textbf{Omni-view} 
& \textbf{Low} 
& \textbf{Automated} 
& \textbf{Scalable} \\
\bottomrule
\end{tabular}%
}
\vspace{-6mm}
\end{table}

To address this gap, we propose \tool, an annotation system specifically designed for multi-camera tracking data. \tool enables annotators to work directly on an omni-view BEV plane: a single ground-plane annotation is automatically projected into 2D bounding boxes across all calibrated camera views, inherently ensuring identity consistency without manual cross-view alignment (Figure~\ref{fig:multi_camera_motivation}b). To further accelerate the labeling process, \tool incorporates two complementary mechanisms: a \textit{Temporal Aware} module that propagates annotations from preceding frames to initialize new ones, and a \textit{Multi-camera Semi-annotation} module that leverages off-the-shelf 2D detectors~\cite{tian2025yolov12} combined with foot-point estimation to automatically generate candidate BEV positions for annotator verification.

We validate \tool through a pilot study on multi-camera broiler tracking and demonstrate substantial reductions in annotation time compared to conventional single-camera workflows. Our contributions are threefold: (1) We present \tool, a multi-camera annotation tool that leverages camera calibration to enable BEV-based labeling, simplifying cross-view annotation and ensuring consistent object identities across all views; (2) We introduce a \textit{Temporal Aware} module and a \textit{Multi-camera Semi-annotation} module that collectively accelerate the annotation process through temporal propagation and automated detection lifting; and (3) We conduct a pilot study on multi-camera broiler tracking, demonstrating that \tool significantly reduces annotation time compared to conventional single-camera labeling tools.

%% file: sec/2_method.tex

\section{ArgusTrack}
\begin{figure*}[t]
\centering
\includegraphics[width=\linewidth]{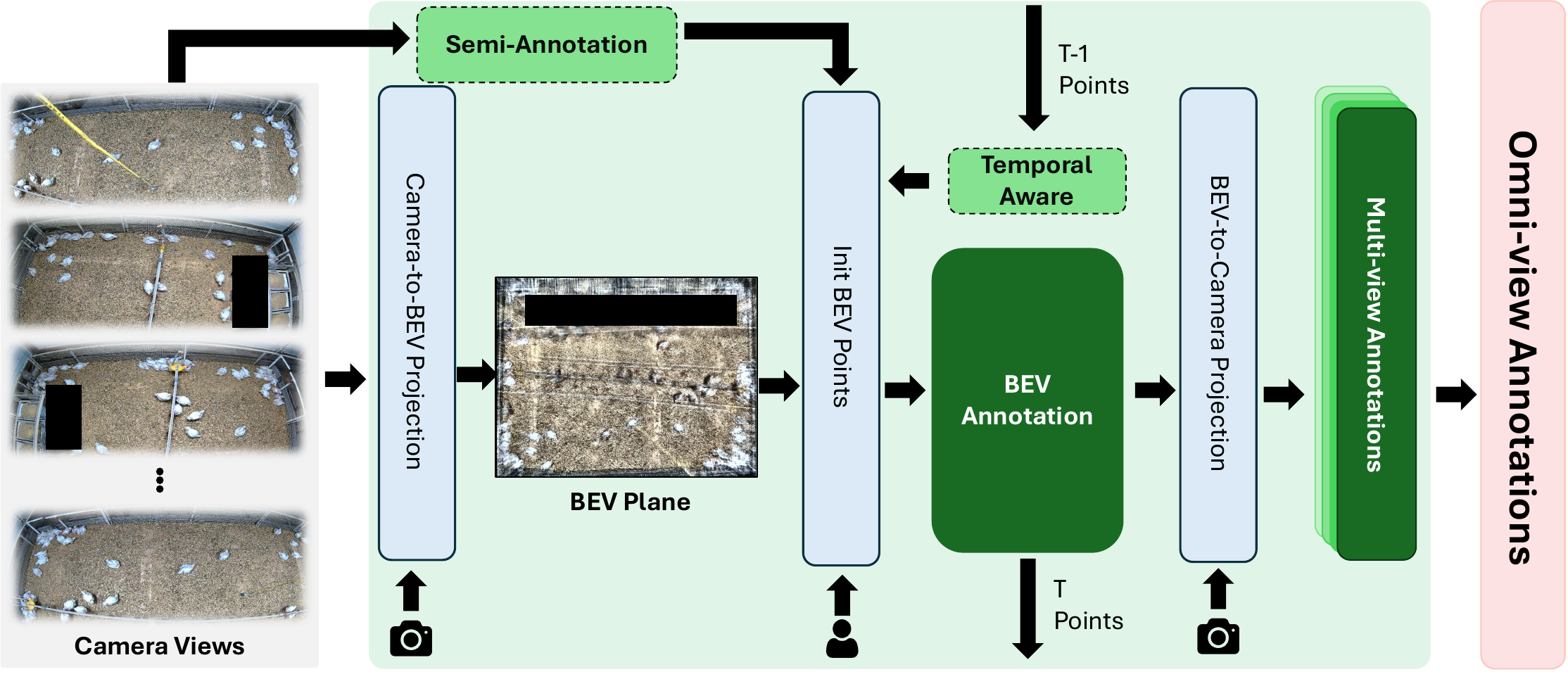}

\vspace{-4mm}
\caption{\textbf{Overview of \tool.} Multi-camera images are projected onto an omni-view BEV plane, where annotators label objects once and the annotations are automatically back-projected to all camera views. The \textit{Temporal Aware} module propagates BEV annotations from frame $t$ to frame $t{+}1$, while the \textit{Semi-annotation} module proposes candidate positions from 2D detections to assist annotator verification.  }
\label{fig:label_tool}
\vspace{-6mm}

\end{figure*}

This section details \tool, as illustrated in Fig.~\ref{fig:label_tool}. We first describe the camera model and BEV projection (Sec.~\ref{sec:camera_model}--\ref{sec:cam_to_bev}), then the BEV-based bounding box generation (Sec.~\ref{sec:bbox}), the Temporal Aware module (Sec.~\ref{sec:temporal}), and finally the Multi-camera Semi-annotation module (Sec.~\ref{sec:semi_annotate}).

\subsection{Camera Model}
\label{sec:camera_model}
A core design principle of \tool is to allow annotators to work on an omni-view BEV plane rather than annotating each camera view independently. We assume a standard pinhole camera model. For each camera $c$, we define the full projection matrix:
\begin{equation}
    \mathbf{H}_c = \mathbf{K}_c \left[ \mathbf{R}_c \mid \mathbf{t}_c \right] \in \mathbb{R}^{3 \times 4},
    \label{eq:proj_matrix}
\end{equation}
where $\mathbf{K}_c \in \mathbb{R}^{3 \times 3}$ is the intrinsic matrix and $[\mathbf{R}_c \mid \mathbf{t}_c] \in \mathbb{R}^{3 \times 4}$ is the extrinsic matrix. The projection from a 3D world point $\mathbf{X}_w = [X, Y, Z, 1]^\top$ to a 2D image point $\mathbf{x} = [u, v, 1]^\top$ is then:
\begin{equation}
    \mathbf{x} \sim \mathbf{H}_c \, \mathbf{X}_w,
    \label{eq:projection}
\end{equation}
where $\sim$ denotes equality up to a projective scale factor.

\subsection{From Camera to BEV}
\label{sec:cam_to_bev}
Each camera image is warped onto the BEV ground plane by inverting the projection in Eq.~\eqref{eq:projection}. Specifically, for ground-plane points ($Z = 0$), $\mathbf{H}_c$ reduces to a $3{\times}3$ homography by discarding the third column, and its inverse $\mathbf{H}_c^{-1}$ maps each BEV pixel $(u, v)$ back to the corresponding image coordinate. Pixel intensities are sampled via bilinear interpolation, and overlapping regions from multiple cameras are blended using per-camera visibility masks, yielding a unified top-down view for annotation.

\subsection{From BEV Point to 2D Bounding Box}
\label{sec:bbox}

On the BEV plane, the annotator places a point $\mathbf{p} = (x, y)$ representing the ground contact point of an object. Each object is associated with a configurable 3D bounding box with dimensions $(w_o, d_o, h_o)$ representing the width, depth, and height of the object in world coordinates. These dimensions can be adjusted per object to accommodate varying object sizes. A default 3D bounding box size is provided based on real-world prior knowledge and can be modified by the annotator as needed.
Given the ground-plane point $\mathbf{p}$ and the object dimensions, the eight corners of the 3D bounding box are defined as:
\begin{equation}
    \mathbf{X}_i = \begin{bmatrix} x + \delta x_i \\ y + \delta y_i \\ \delta z_i \\ 1 \end{bmatrix}, \quad i = 1, \dots, 8,
\end{equation}
where $(\delta x_i, \delta y_i, \delta z_i) \in \left\{ \pm \frac{w_o}{2} \right\} \times \left\{ \pm \frac{d_o}{2} \right\} \times \left\{ 0, h_o \right\}$ enumerate all corner offsets relative to the ground-plane center.
Each corner is then projected into camera $c$ using Eq.~\eqref{eq:projection}

The 2D bounding box in camera $c$ is obtained by computing the axis-aligned enclosing rectangle of all projected corners: \begin{equation} \text{BBox}^c = \left( \min_i u_i^c,\; \min_i v_i^c,\; \max_i u_i^c,\; \max_i v_i^c \right), \end{equation} where $(u_i^c, v_i^c)$ are the projected image coordinates of corner $i$. Note that in the annotation interface, only the BEV point and the resulting 2D bounding boxes are displayed; the intermediate 3D bounding box is not visualized to keep the interface clean and intuitive.

\subsection{Temporal Aware}
\label{sec:temporal}
To improve annotation efficiency, the tool supports propagating bounding boxes from the previous frame as initialization for the current frame. When a new frame is loaded, users can copy bounding boxes from the preceding frame, preserving both positions and identities. Since consecutive frames typically exhibit only small positional changes, annotators need only minor adjustments to align boxes with current object positions. This feature is optional and can be skipped in cases such as scene cuts or significant camera motion.

\subsection{Multi-camera Semi-annotation}
\label{sec:semi_annotate}
\begin{figure*}
    \centering
    \includegraphics[width=1\linewidth]{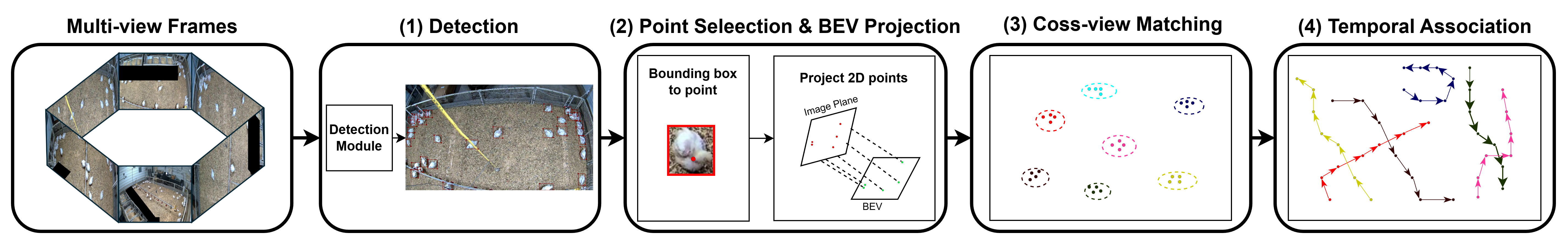}
\vspace{-6mm}
    
    \caption{\textbf{Semi-Annotation Module.} The pipeline consists of four stages: (1) \textbf{Detection}: a pretrained detector generates 2D bounding boxes in each camera view; (2) \textbf{Point Selection \& BEV Projection}: foot-point estimation and homography projection project detections onto the BEV plane; (3) \textbf{Cross-view Matching}: groups BEV points into clusters to assign consistent identities across cameras; (4) \textbf{Temporal Association}: links identities across frames.}
    \label{fig:semi_annotate}
\vspace{-6mm}

\end{figure*}

This section describes the semi-annotation module, which leverages a pretrained object detection model to automatically generate initial labels, reducing manual annotation effort. Annotators only need to review, refine, and supplement missed detections. The overall architecture is illustrated in Figure~\ref{fig:semi_annotate}, comprising four key components: (1) \textbf{Detection}, (2) \textbf{Point Selection and BEV Projection}, (3) \textbf{Cross-view Matching}, and (4) \textbf{Temporal Association}.

\subsubsection{Detection Module}
A pretrained YOLO-based model~\cite{wang2023yolov7,varghese2024yolov8,tian2025yolov12} automatically generates initial bounding box labels for each frame. Each detected bounding box, along with its class label and confidence score, is forwarded to subsequent stages.

\subsubsection{Point Selection and BEV Projection}
For each high-confidence detection, the projection point is a weighted interpolation between the bounding box center $p_{\text{center}}$ and bottom-center $p_{\text{bottom}}$:
\begin{equation}
    p = (1 - t) \cdot p_{\text{center}} + t \cdot p_{\text{bottom}},
    \label{eq:proj_point}
\end{equation}
where $t \in [0,1]$ is adaptively determined based on camera-to-object distance and elevation angle. The point $p = [u, v]^\top$ is then lifted to homogeneous coordinates and mapped onto the BEV ground plane via $\mathbf{H}_c$ (Eq.~\eqref{eq:proj_matrix}).

\subsubsection{Cross-view Matching}
BEV projections from all cameras are clustered using K-Means, where each cluster corresponds to a single real-world object. The number of clusters $K$ is user-defined. A constrained greedy assignment enforces at most one detection per camera per cluster, with previous-frame centroids used to warm-start initialization. The output maps each \textit{(camera, local\_id)} to a consistent \textit{global\_id}.
\subsubsection{Temporal Association}
To maintain stable tracks across frames, the Hungarian algorithm is applied on a cost matrix $\mathbf{C} \in \mathbb{R}^{K \times M_f}$, where $K$ is the number of tracks and $M_f$ is the number of detections in the current frame. Each entry represents the BEV Euclidean distance between a track's last known position $p_t$ and a detection $p_d$:
\vspace{-2mm}
\begin{equation}
    \mathbf{C}[t, d] = \left\| p_t - p_d \right\|_2
    \label{eq:cost_matrix}
\end{equation}
A match is accepted only if $\mathbf{C}[t,d] \leq \delta$. Unmatched tracks retain their previous positions until reassigned, ensuring consistent global identities throughout the sequence.
\vspace{-2mm}

%% file: sec/3_exp.tex
\section{Experiment}

To evaluate the effectiveness of \tool, we conduct a user study comparing annotation speed against CVAT~\cite{CVAT_ai_Corporation_Computer_Vision_Annotation_2023}, a widely used open-source 2D labeling tool. We select 20 continuous timestamp scenes containing 11 synchronized camera views of broilers in a poultry house. Annotators are asked to annotate all visible broilers across all camera views for each scene, assigning consistent tracking IDs across views. We measure the total annotation time and report the average time per scene.

\subsection{Annotation Speed Comparison}

\begin{table}[t]
\centering
\caption{Average time per scene (in seconds) between CVAT and \tool.}
\vspace{-2mm}
\label{tab:speed_comparison}
\resizebox{0.9\linewidth}{!}{%
\begin{tabular}{l c c}
\toprule
\textbf{Tool} & \textbf{Avg. Time / Scene (s)} & \textbf{Speedup} \\
\midrule
CVAT~\cite{CVAT_ai_Corporation_Computer_Vision_Annotation_2023} & 1872 & $1.0\times$ \\
TrackMe~\cite{phan2024trackme} & 1392 & $1.34\times$ \\
\tool (Ours) & 102 & $18.35\times$ \\
\bottomrule
\end{tabular}%
}
\vspace{-4mm}
\end{table}

Table~\ref{tab:speed_comparison} compares the annotation time between CVAT and \tool. With conventional single-camera tools, annotators must label each of the $C$ camera views independently for $N$ objects, resulting in $\mathcal{O}(N \times C)$ effort, plus additional time to verify identity consistency across views---a process prone to errors when objects are visually similar. In contrast, \tool reduces per-scene effort to $\mathcal{O}(N)$ by enabling omni-view BEV annotation with automatic cross-view projection, achieving an $18.35\times$ speedup over CVAT.
\subsection{Ablation Study}

\begin{table}[t]
\centering
\caption{Effect of progressively adding modules to the base \tool pipeline.}
\vspace{-2mm}

\label{tab:ablation}
\resizebox{0.6\linewidth}{!}{%
\begin{tabular}{l|c}
\toprule
\textbf{Method}  & \textbf{Time (s)} \\
\midrule
\tool (Base) & 150 \\
+ TA & 113 \\
+ SA  & 102 \\
\bottomrule
\end{tabular}%
}
\vspace{-4mm}

\end{table}

We conduct an ablation study to quantify the contribution of each proposed module, with results reported in Table~\ref{tab:ablation}. Starting from the base \tool pipeline (omni-view BEV annotation with automatic projection), we progressively enable the Temporal Aware (TA) and Semi-Annotation (SA) modules. The base pipeline already provides substantial improvement over single-camera workflows by eliminating redundant cross-view labeling. Adding the TA module reduces annotation time from 150s to 113s by propagating annotations from preceding frames, allowing annotators to adjust existing labels rather than drawing from scratch. The SA module further reduces time to 102s by automatically generating candidate BEV positions from 2D detections, shifting the annotator's task from creation to verification. Together, both modules contribute measurable and complementary time savings on top of the base pipeline.
\vspace{-2mm}

\section{Conclusion}

In this paper, we presented \tool, an efficient multi-camera annotation tool that leverages camera calibration to enable BEV-based labeling with automatic cross-view projection, reducing per-frame effort from $\mathcal{O}(N \times C)$ to $\mathcal{O}(N)$. Two complementary modules further accelerate the process: \textit{Temporal Aware} propagation and \textit{Multi-camera Semi-annotation} via off-the-shelf 2D detectors. Our pilot study confirms that \tool significantly reduces annotation time compared to conventional tools such as CVAT.